# TrialCalibre: A Fully Automated Causal Engine for RCT Benchmarking and Observational Trial Calibration

Amir Habibdoust [1]  Xing Song [2]
ICML 2025

## Abstract

Real-world evidence (RWE) studies that emulate target trials increasingly inform regulatory and clinical decisions, yet residual, hard-to-quantify biases still limit their credibility. The recently proposed BenchExCal framework addresses this challenge via a two-stage Benchmark, Expand, Calibrate process, which first compares an observational emulation against an existing randomized controlled trial (RCT), then uses observed divergence to calibrate a second emulation for a new indication causal effect estimation. While methodologically powerful, BenchExCal is resource-intensive and difficult to scale. We introduce TrialCalibre, a conceptualized multi-agent system designed to automate and scale the BenchExCal workflow. Our framework features specialized agents—such as the Orchestrator, Protocol Design, Data Synthesis, Clinical Validation, and Quantitative Calibration Agents—that coordinate the the overall process. TrialCalibre incorporates agent learning (e.g., RLHF) and knowledge blackboards to support adaptive, auditable, and transparent causal effect estimation.

## 1. Introduction

The use of Real-World Evidence (RWE) or causal effect esitmation derived from Real-World Data (RWD) is transforming medicine and pharmaceutical research (Costa et al., 2025; Zhu et al., 2023; Vaghela et al., 2024; Burns et al., 2023).A key application is supporting the expansion of indications for existing therapeutic products or treatment, which can expedite patient access to beneficial treatments. Target Trial Emulation (TTE), a methodology for designing observational studies to explicitly mimic a randomized controlled trial (RCT), provides a strong foundation for generating robust RWE(Hernán & Robins, 2016; Danaei et al., 2018).Wang et al. recently proposed the "Benchmark, Expand, and Calibration" (BenchExCal) approach(Wang et al., 2025a). BenchExCal is an advanced, two-stage trial emulation strategy. In Stage 1, a database study emulates a completed RCT for an existing indication of a drug, and its results are benchmarked against the RCT to quantify any "divergence." In Stage 2, learnings from Stage 1 inform the emulation of a hypothetical trial for a new, expanded indication, and the results are "calibrated" using the divergence observed in Stage 1. This calibration aims to account for systematic differences between RCTs and causal inference studies, thereby improving the confidence in the RWE findings for the new indication(Wang et al., 2025a). While conceptually powerful, the BenchExCal process involves complex steps: selection of appropriate RCTs for benchmarking, meticulous emulation design for two separate target trials, quantification of divergence, appropriate use of the divergence measurement to inform calibration, and interpretation. These steps require extensive and multidisciplinary domain expertise, which can be resource intensive and can hinder the scalability of the approach.

Concurrently, advancements in artificial intelligence (AI), particularly multi-agent systems (MAS), have demonstrated the potential to automate complex scientific workflows. While there are several research attempts to fully or partially automate the causal inference process(Costa et al., 2025; Zhu et al., 2023; Vaghela et al., 2024; Burns et al., 2023; Hernan & Robins, 2016; Danaei et al., 2018; Wang et al., 2025a; Le et al., 2024; Khatibi et al., 2025; Vashishtha et al., 2023; Biza et al., 2025; Wang et al., 2025b), applications to the health domain is limited. For instance,Gonzalez et al. proposed TRIALSCOPE, a unifying causal framework that leverages biomedical language models to scale real-world evidence generation and trial emulation efforts(Gonzalez et al., 2023). Li et al. introduced TrialGenie, an agentic AI framework to automate various components of TTE, from protocol parsing to cohort generation and statistical analy-sis(Li et al., 2025). This paper introduces "TrialCalibre", a conceptual framework for automating the BenchExCal workflow: from initial benchmarking to final calibrated evidence generation. Our aim is to articulate the vision and outline a viable system architect with core components for such an automated agentic framework, to catalyze discussions around its feasibility and potential impact.

[1]Institute for Data Science and Informatics, University of Missouri, Columbia, MO, USA. [2]Department of Biomedical Informatics, Biostatistics, and Medical Epidemiology, University of Missouri, Columbia, MO, USA. Correspondence to: Amir Habibdoust <ahb3b@missouri.edu>.

## 2. The BenchExCal : A Brief Overview

The BenchExCal approach, as detailed by Wang et al.(Wang et al., 2025a) , consists of three main steps across two stages:

### 2.1. Stage 1: Benchmark

1. An initial database study ($RWE_1$) is designed to emulate a completed RCT ($RCT_1$) for an existing indication of the drug of interest.

2. The treatment effect estimate from $RWE_1$ ($\hat{\theta}_1^*$) is compared to the observed effect from $RCT_1$ ($\hat{\theta}_1$).

3. The divergence, $\hat{\xi}_1 = \hat{\theta}_1^* - \hat{\theta}_1$, is quantified. This divergence reflects the net difference due to residual confounding, misclassification, population differences, and other systematic variations between the RCT and its RWD emulation.

### 2.2. Stage 2: Expand and Calibrate

1. Expand: Using the established design, data, and analytical methods from Stage 1, a second database study ($RWE_2$) is conducted to emulate a hypothetical target trial ($RCT_2$) for a new or expanded indication. This yields an effect estimate $\hat{\theta}_2^*$.

2. Calibrate: The divergence observed in Stage 1 ($\hat{\xi}_1$) is used to inform the interpretation of $\hat{\theta}_2^*$. This typically involves scaling $\xi_1$ to an appropriate value for Stage 2 ($\xi_2$) and incorporating it into sensitivity analyses (e.g., Bayesian methods where $\xi_2$ informs a prior distribution) to produce a calibrated estimate or range for the treatment effect in the expanded indication.

## 3. Conceptual Framework for TrialCalibre

TrialCalibre is conceptualized as a Hybrid Hierarchical-Blackboard MAS. This architecture leverages a hierarchical structure for overall workflow management and specialized blackboard systems for collaborative, knowledge-intensive tasks. All communication and task orchestration are handled by a coordinated multi-agent framework. This relies on structured exchanges managed by the central Orchestrator Agent, with agents collaborating through shared knowledge blackboards and direct communication channels, ensuring reliable, auditable, and scalable interactions (Figure 1.)

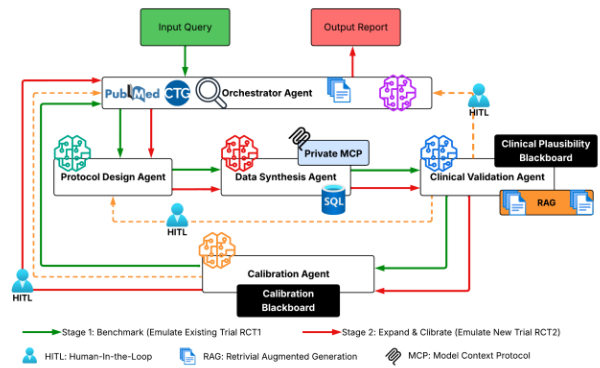

Figure 1. High-Level Conceptual Workflow of TrialCalibre.

Orchestrator Agent: This Agent manages the end-to-end workflow across Stage 1 (Benchmark) and Stage 2 (Expand & Calibrate). It coordinates agent activities by initiating specific tasks, synthesizing outputs, and compiling final reports. A critical function is its Stage 1 concordance assessment—based on inputs from the Clinical Validation and Quantitative Calibration Agents, with potential Human-in-the-Loop (HITL) checkpoints—which determines whether the workflow progresses to Stage 2. Upon receiving a user query, the Orchestrator Agent leverages advanced large language model (LLM) capabilities for natural language understanding and planning to decompose the request. If new evidence generation via the BenchExCal approach is warranted, it initiates and coordinates a preliminary investigation to identify suitable candidate benchmark trials. This involves orchestrating literature and registry searches and collaborating closely with the Clinical Validation Agent to assess clinical relevance before Stage 1 commences.

Protocol Design Agent: It has crucial role in causal study design. The agent operates in two distinct capacities, corresponding to the two stages. In Stage 1, it retrieves and design standardizes the existing randomized controlled trial protocol used for benchmarking purposes (replicate $RCT_1$). It uses data from Data Synthesis Agent & Clinical Validation Agent to design correct protocols. During Stage 2, it facilitates the definition of a protocol for the hypothetical trial ($RCT_2$) tailored to a new indication, ensuring both alignment with the benchmark and methodological clarity.

Data Synthesis Agent: It consistently translates trial protocols into executable queries compatible with RWD sources throughout both Stage 1 and Stage 2. Its responsibilities include managing data cleaning, the construction of patient cohorts, data extraction, mapping variables between data sources, and performing data quality assurance checks. It ensures that methodological consistency is maintained across the $RWE_1$ (estimated causal effect in Stage 1) and $RWE_2$ (Stage 2) datasets. We can leverage private model context protocol to enable agent's secure access to local data.

Clinical Validation Agent: The agent provides domain expertise throughout the BenchExCal workflow, supported by Retrieval-Augmented Generation (RAG). Its key responsibilities include advising on appropriate covariate selection for $RWE_1$ (Stage 1) and $RWE_2$ (Stage 2), and assisting in the clinical interpretation of any observed divergence ($\hat{\xi}_1$) in Stage 1. For Stage 2, it critically evaluates the clinical plausibility of transferring divergence characteristics from the benchmark, potentially employing a "Clinical Plausibility Blackboard" for complex collaborative reasoning to refine judgments. Additionally, during the initial evidence scoping phase managed by the Orchestrator, this agent plays a crucial role in evaluating the clinical relevance, methodological soundness, and suitability of potential benchmark trials (RCTs) by using its domain knowledge and RAG capabilities to assess literature sources.

Quantitative Calibration Agent: Serving as the analytical core of TrialCalibre, the agent is tasked with conducting the formal causal analysis in target trial emulation (TTE) framework. In Stage 1, it carries out the TTE for $RWE_1$, estimates the corresponding treatment effect ($\hat{\theta}_1$), and compares this estimate against the known treatment effect from $RCT_1$ ($\theta_1$). By doing so, it quantifies the divergence ($\xi_1$) along with its uncertainty and conducts concordance checks. In Stage 2, this agent performs the TTE for $RWE_2$ to estimate $\theta_2$ and leverages the Automated Calibration Engine to scale the previously observed divergence $\xi_1$ to a new divergence $\xi_2$. It applies this scaled divergence in sensitivity or Bayesian analyses to produce both uncalibrated and calibrated results. Additionally, it may utilize a "Calibration Blackboard" to systematically explore and select optimal methodologies for divergence scaling or strategies for prior elicitation.

## 4. Key Automated Processes and Innovations

Automating BenchExCal within TrialCalibre involves these key processes, enhanced by the architecture. While Figure 1 provides an overview, Figure 2 details the system's internal communication, agent-specific tasks, and coordination that enable TrialCalibre's autonomous operation:

### 4.1. Intelligent Trial Discovery and Validation

This crucial initial stage of the TrialCalibre workflow. It is activated when a benchmark trial ($RCT_1$) is not predefined by the user. In this phase, Orchestrator initiates the discovery process. Leveraging advanced LLM capabilities, it systematically searches related literature and clinical trial registries for potential candidate RCTs. Subsequently, the Clinical Validation evaluates these trials for their clinical relevance, methodological soundness, and overall suitability as a robust benchmark, employing its domain expertise and RAG tools. This collaborative assessment, guided by the Orchestrator and potentially involving HITL input, ensures the rigorous selection of the most appropriate available $RCT_1$. Successfully completing Stage 0 establishes an automated and evidence-based foundation for the Stage 1 and Stage 2 of the process(See Figure 2). For example, the user asks, "What are the comparative effectivenesses of four different classes of antihypertensive drugs?" The agent, in cooperation with the Clinical Validation, realized that there is no existing trial that directly answers this causal question. However, there are trials that compare two of the drugs—for instance, the ONTARGET trial(Investigators, 2008) —which can be extended to infer an answer.

### 4.2. Automated Stage 1 Benchmarking and Divergence

The Quantitative Calibration would automatically execute the analysis plan for $RWE_1$, compare results with $RCT_1$, compute concordance metrics, and calculate the divergence ($\hat{\xi}_1$) and its variance. These outputs, communicated via the framework's messaging channels, inform the Orchestrator and Clinical Validation for progression decisions. For example, if the divergence of the treatment effect between the original ONTARGET and the replicated ONTARGET is acceptable, they decide to proceed to Stage 2.

### 4.3. Automated Calibration Engine

Operating within the Quantitative Calibration Agent, the Automated Calibration Engine automatically calculates the scaled divergence ($\hat{\xi}_2$) and its uncertainty, potentially leveraging a blackboard for methodological exploration. It generates priors for $\xi_2$ to support Bayesian analyses or defines ranges for tipping-point analyses based on $\hat{\xi}_2$ and its confidence interval. Finally, it performs Bayesian adjustments or tipping-point analyses to execute the calibration.

### 4.4. Iterative Validation, Learning, and Refinement

The validity of BenchExCal depends critically on the similarity of divergence mechanisms between stages. The Clinical Validation Agent evaluates this assumption using Retrieval-Augmented Generation (RAG), its knowledge base, and potentially its "Clinical Plausibility Blackboard." If transferability is uncertain, the Orchestrator, informed by the inter-agent communication system, may halt or flag Stage 2 results. To enhance these critical judgments, RLHF—potentially augmented by causal reinforcement learning (Zhang & Bareinboim, 2020; Blübaum & Heindorf, 2024) —is integral. Experts provide feedback (e.g., ratings) on agent decisions regarding divergence assessment, transferability, and calibration parameters. This feedback, processed through RLHF methods, iteratively refines the decision-making policies of the Clinical Validation, Quanti-tative Calibration, and Orchestrator, aligning outputs with expert intuition. The framework's design, particularly the logging of agent interactions and data flow, ensures feedback, agent actions, and outcomes are systematically logged and traceable, supporting robust learning.

4.5. Structured Reporting:

TrialCalibre generates reports for Stage 1 (benchmarking, divergence $\xi_1$) and Stage 2 (uncalibrated $RWE_2$ results, calibration process, calibrated findings, tipping-point analyses). The inherent observability of agent actions and data flow within this system will enhance the transparency and repro-ducibility in these reports.(Hansford et al., 2023)

4.6. Automated Adaptive Protocol Generation:

A key innovation within TrialCalibre is the automated and adaptive generation of emulation protocols, which is critical to ensuring the robustness and real-world applicability of the Stage 1 benchmark. The process begins with the Protocol Design retrieving and standardizing the historical protocol as an initial blueprint. The blueprint is then iteratively refined through a collaborative process orchestrated by the Orchestrator, using coordinated communication: the Data Synthesis assesses it against available RWD to identify limitations and potential proxy variables, while the Clinical Validation provides guidance on the clinical acceptability of those proxies. The Protocol Design then finalizes an adapted emulation protocol for ($RCT_1$), transparently documenting all deviations and justifications. This ensures a feasible and methodologically sound foundation for divergence quantification and subsequently informs the definition of Stage 2 protocol. While TrialCalibre builds on common multi-agent LLM tools like retrieval and prompting, its novelty lies in applying them to causal benchmarking, with dedicated audit agents and integration into the workflow. It uniquely incorporates counterfactual checks and RCT-based validation.

## 5. Challenges

The successful adoption of TrialCalibre depends on overcoming several key challenges. One major hurdle is the limitation of RWD, which often lacks granular clinical details (e.g., blood pressure in claims). This requires protocol adaptation by the Protocol Design and Data Synthesis Agents, possibly using proxies or noting data gaps, with the Clinical Validation Agent evaluating the impact on study validity. Enhancing LLM Causal Reasoning is another challenge. To enhance LLM understanding of the three levels of Pearl's causal ladder, models must advance beyond RAG. Specifically, integrating structured causal knowledge via domain-specific Directed Acyclic Graphs (DAGs) can significantly improve the causal reasoning ability of LLMs, helping them distinguish causation from correlation. Automating complex clinical decisions—such as assessing divergence transferability or selecting calibration parameters—and designing effective, unbiased RLHF for decision-making agents (e.g., Clinical Validation, Quantitative Calibration, Orchestrator) remain open research problems. Strong human-AI collaboration is essential for clinical validity. Another frontier lies in advancing and validating causal reasoning in TrialCalibre's LLM-powered agents, developing appropriate evaluation metrics, and applying this reasoning to improve decisions via methods like Causal Reinforcement Learning. While privacy is always a consideration in healthcare AI, TrialCalibre is designed to operate on de-identified data, and no major privacy concerns are anticipated at this stage. Fi-nally, ensuring the methodological robustness of automated calibration—especially by the Quantitative Agent using lim-ited or noisy benchmarks—is critical. Building stakeholder trust will require transparent operations, auditable decision trails, and outputs that are both reliable and interpretable. TrialCalibre's multi-agent setup requires computational re-sources, but remains more efficient than manual alternatives. Efficiency can be improved through smaller models, early stopping, and reuse of shared steps.

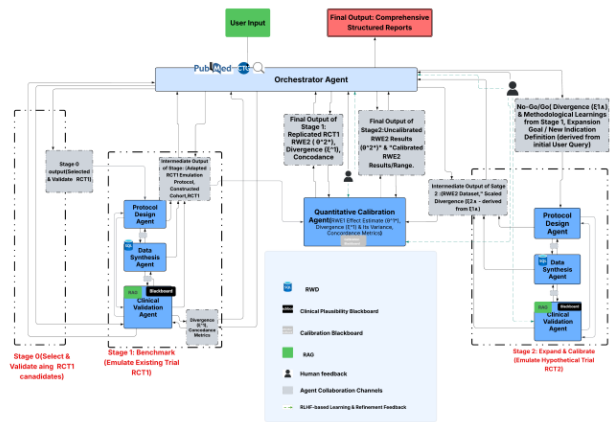

Figure 2. Detailed Architecture of TrialCalibre Interactions and Data Flow. Intelligent benchmark trial discovery (Phase 0), Stage 1 (Replicating an existing RCT1 with adaptive protocol design and divergence quantification), and Stage 2 (Expand & Calibrate), all coordinated by a central Orchestrator with specialized agent collaboration, communication, and RLHF-driven agent learning.

## 6. Conclusion

The BenchExCal framework is a pivotal advancement for RWE and causal effect estimation in indication expansion. By utilizing agentic AI, the proposed TrialCalibre aims to make this methodology more scalable, efficient, and transparent. Addressing the inherent complexities through intelligent automation and adaptive learning via RLHF presents a compelling path to accelerate high-quality RWE generation, ultimately benefiting patient care by facilitating more timely and robust evidence for new therapeutic indications.

## Impact Statement

This work introduces TrialCalibre, an automated system to advance real-world evidence (RWE) generation for therapeutic indication expansion, aiming to accelerate patient access to treatments and improve the robustness of RWE for critical medical decisions. While this research contributes to the field of Machine Learning, its direct application in healthcare necessitates careful consideration of societal impacts. Key among these are ensuring the system's accuracy, transparency, and fairness to protect patient safety and foster trustworthy clinical application. Consequently, responsible development, alongside robust governance frameworks, will be vital for realizing its beneficial societal impact.

## Acknowledgements

We are grateful to the anonymous reviewers for their insightful comments, which helped improve this manuscript.